# Anchor-free Small-scale Multispectral Pedestrian Detection

Alexander Wolpert[1,2]
alexander.wolpert@student.kit.edu

Michael Teutsch[2]
michael.teutsch@hensoldt.net

M. Saquib Sarfraz[1,3]
saquib.sarfraz@kit.edu

Rainer Stiefelhagen[1]
rainer.stiefelhagen@kit.edu

[1] Institute of Anthropomatics & Robotics
Karlsruhe Institute of Technology
Karlsruhe, Germany

[2] Hensoldt Optronics GmbH
Oberkochen, Germany

[3] Autonomous Systems
Daimler TSS
Germany

**Abstract**

Multispectral images consisting of aligned visual-optical (VIS) and thermal infrared (IR) image pairs are well-suited for practical applications like autonomous driving or visual surveillance. Such data can be used to increase the performance of pedestrian detection especially for weakly illuminated, small-scaled, or partially occluded instances. The current state-of-the-art is based on variants of Faster R-CNN and thus passes through two stages: a proposal generator network with handcrafted anchor boxes for object localization and a classification network for verifying the object category. In this paper we propose a method for effective and efficient multispectral fusion of the two modalities in an adapted single-stage anchor-free base architecture. We aim at learning pedestrian representations based on object center and scale rather than direct bounding box predictions. In this way, we can both simplify the network architecture and achieve higher detection performance, especially for pedestrians under occlusion or at low object resolution. In addition, we provide a study on well-suited multispectral data augmentation techniques that improve the commonly used augmentations. The results show our method's effectiveness in detecting small-scaled pedestrians. We achieve 5.68 % log-average miss rate in comparison to the best current state-of-the-art of 7.49 % (∼25 % improvement) on the challenging KAIST Multispectral Pedestrian Detection Benchmark.

## 1 Introduction

Pedestrian detection is an important research topic in the field of computer vision. It is of practical use to increase the level of automation in many applications such as robotics, autonomous vehicles, visual surveillance, or search and rescue. Admittedly great progress was made on pedestrian detection in visual-optical (VIS) color images over the years [57]. As good as those approaches perform on clear daylight images, their performance drastically deteriorates under challenging conditions such as adverse lighting conditions, occlusions, or low object resolution. It has since been established that adopting additional, complementary modalities can improve the performance under such difficult conditions [14, 38, 39]. One exceptionally well-suited additional modality are thermal infrared (IR) images: since humans often have a higher temperature compared to the surrounding background, their emitted radiation can be sensed well by thermal IR cameras [17]. The appearing bright spots in IR images together with the aligned texture patterns in the related VIS images can guide and support detection approaches especially under the mentioned challenging conditions. This was impressively demonstrated on the publicly available KAIST Multispectral Pedestrian Detection Benchmark [24]. As a result, multispectral pedestrian detection became a rising field of research in recent years [28, 32, 35, 55]. All current state-of-the-art methods, however, have something in common: they are variants of the popular deep learning based object detection approach Faster R-CNN [42]. Detection is performed in two stages using a proposal generator called Region Proposal Network (RPN) for object localization, followed by a classification network to determine the object class. Localization is based on a set of handcrafted anchor boxes that are regressed towards the ground truth (GT) bounding boxes during training.

Just recently, novel anchor-free object detection methods were proposed in the VIS spectrum [36, 37, 43] that focus on learning pedestrian representations based on object center and scale instead of direct bounding box predictions. This approach performs particularly well for small-scaled or occluded pedestrians since the object center provides more distinctive information than the object boundaries. Furthermore, this method is not only more effective with a higher detection rate but also more efficient as it is performed within a single stage. In this paper, we adopt this idea and propose a novel multispectral fusion approach based on the Center and Scale Prediction Network (CSPNet) [36, 37]. In contrast to prior literature, we fuse feature maps of individual VIS and IR Deep Convolutional Neural Networks (DCNNs) after multiple convolution layers creating rich hierarchies of multispectral feature maps that are concatenated and evaluated by only one detection head. To the best of our knowledge, this is the first anchor-free approach for multispectral pedestrian detection.

Our contribution is fourfold: (1) we propose and analyze multiple different fusion techniques based on the CSPNet architecture, (2) we propose a novel multispectral data augmentation and introduce a general innovative approach for data augmentation on multispectral images, (3) we analyze the influence and the impact of the three currently available training annotations [24, 32, 55] on the small-scale pedestrian instances of the KAIST dataset, and (4) we demonstrate the effectiveness of the proposed method by improving the current state-of-the-art on the KAIST benchmark by about 23 % for small-scale pedestrian instances with a height of 40 pixels or less and by about 25 % on the KAIST *reasonable* subset that contains well visible pedestrian instances only.

The remainder of this paper is organized as follows: related work is reviewed in Section 2. The proposed multispectral fusion approach is described in Section 3. Experimental results are presented in Section 4. We conclude in Section 5.

## 2 Related Work

**Multispectral Pedestrian Detection:** Automatic pedestrian detection has been an active field of research in computer vision for many years [4, 8, 13, 51]. Most efforts are made for VIS camera data [1, 12, 23, 56]. But since thermal IR cameras became more affordable [17], pedestrian detection in the IR spectrum is a continuously rising research topic especially for surveillance [2, 22, 49] but also for automotive applications [11, 15, 18]. A former niche topic attracting more and more attention in recent years is multispectral pedestrian detection: color VIS and thermal IR images are acquired simultaneously and aligned to generate four or six channel images. The core idea is to favor the VIS spectrum under well illuminated daytime conditions and the thermal IR spectrum in case of large relative temperature differences within the scene [28, 35]. Early publications on multispectral person detection appeared with the publication of the OSU Color-Thermal Database [9, 10]. Proposals are generated using background subtraction and then evaluated using contour based fusion [10], periodic gait analysis [31], active contours [54], or Riemannian manifolds [45]. A new impulse was set with the introduction of the KAIST Multispectral Pedestrian Benchmark [24]. First benchmark results were reported using a combination of Aggregated Channel Features (ACF) and Boosted Forest (BF) [24]. Wagner et al. [52] added a multispectral fusion DCNN to further improve classification. Inspired by Faster R-CNN [42], multiple approaches for multispectral fusion DCNNs followed, that usually fuse convolutional feature maps of individual VIS and IR DCNNs halfway [5, 16, 28, 32, 35, 40]. Further improvement at the price of more complex DCNN architectures was achieved by introducing illumination-aware weighting functions within a late fusion strategy [19, 33]. Currently, the state-of-the-art results are given by approaches inspired by Faster R-CNN as well: Multispectral Simultaneous Detection and Segmentation R-CNN (MSDS-RCNN) [32] benefits from multi-task learning using semantic instance segmentation [3] and Aligned Region CNN (AR-CNN) [55] implicitly compensates for the multispectral misalignment error.

**Anchor-free Object Detection:** In contrast to most recent object detection methods, *anchor-free* or *box-free* approaches do not need anchor boxes for object localization. This is an advantage since handcrafting anchor boxes requires prior knowledge about the objects of interest. Anchor-free approaches were first introduced as proposal generators (first stage) within two-stage detection approaches regressing to bounding box corners [50] or center points of extremely small-scaled objects [29]. They, however, can be used as efficient single-stage detectors as well: Topological Line Localization (TTL) [48] predicts a heatmap for the pedestrians' top and bottom vertices from the upsampled intermediate feature maps of a ResNet-50 backbone. CornerNet [30] follows a similar approach with predicting heatmaps that contain an object's top-left and bottom-right corners as well as an embedding vector that defines the pairing between both corners. Most recently, CSPNet [36, 37] improved the state-of-the-art in pedestrian detection significantly. The architecture is similar to TLL. However, only a single object point is predicted via an object center heatmap together with each object's scale. Furthermore, complex post-processing such as Markov Random Fields (MRFs) can be omitted. Many anchor-less approaches use pixel-wise loss functions. To achieve a smoothly converging loss, they supplement the GT points using Gaussian distributions. As an alternative to this approach, [43] proposed a novel loss called weighted Hausdorff distance to predict object centers directly.

## 3 Proposed Method

### 3.1 Multispectral Box-less Pedestrian Detection

Anchor box-less detectors recently became the state-of-the-art for many object detection tasks. They are thus a natural choice to be investigated as a backbone for multispectral pedestrian detection. Prior to this work, we investigated different architectures [37, 43, 48] and found that the recently proposed CSPNet [37] stands out as it manages to combine the slim architecture of a single-stage detector while at the same time it avoids tedious post-processing in form of MRFs or embedding vectors [30, 48].

The CSPNet consists of two modules. In the first module, a ResNet-50 backbone is used to extract features from a color image with size $(H \times W)$ via five convolutional layers. All intermediate feature maps from layers three to five are cached, normalized, and upsampled to a resolution of $(\frac{H}{4} \times \frac{W}{4})$. Earlier feature maps are not included since the authors showed that this does not increase the detection performance but the inference time and the number of trainable parameters instead. The second module is the detection head. There, the upscaled feature maps taken from the three convolutional layers are concatenated and go through a final convolutional layer. Its output is passed to three separate prediction layers to calculate (1) the center heatmap, (2) the scale map, and (3) the offset map using a combination of regression losses and a binary cross-entropy loss adopted with focal loss. Since the feature maps are of resolution $(\frac{H}{4} \times \frac{W}{4})$ the entries of the offset map are added to the upsampled center coordinates to counter numerical inaccuracies. Finally, given center, scale, and offset of a pedestrian, a related bounding box can be determined. For an in-depth description of the CSPNet, please refer to the original paper [37]. However, CSPNet is not designed to process multispectral data. Our contribution is to modify and adapt it by introducing specialized multispectral fusion blocks within its architecture.

## 3.2 Proposed Multispectral Fusion

The established approach for fusion of color VIS and thermal IR data in the field of multispectral pedestrian detection is to have two feature extraction streams that process each modality separately. At some layer, both streams are merged in the feature extraction module and the fused data is processed further [28, 32, 35, 52].

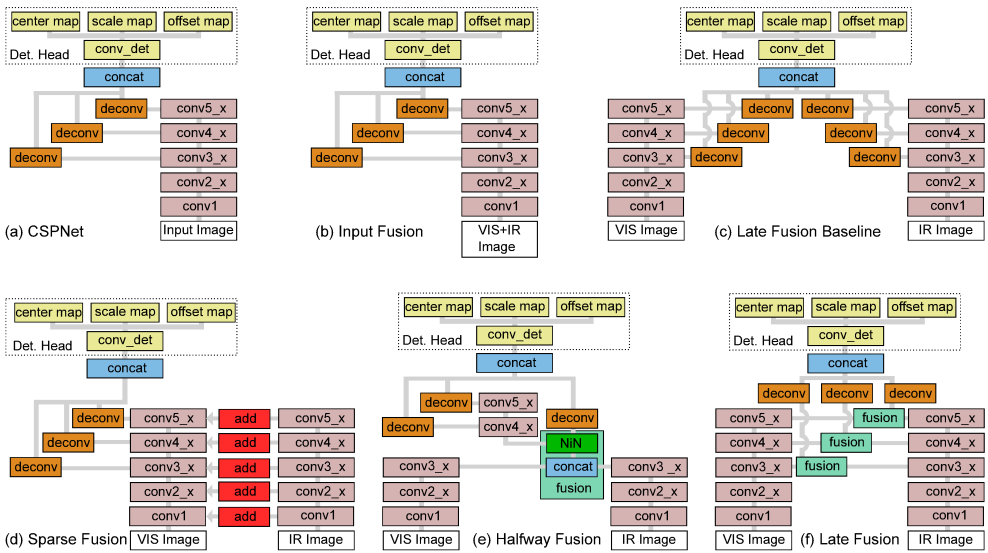

Figure 1: An overview of the baseline architecture (a) and the considered five fusion architectures (b) - (f). The *fusion* blocks (green) are implemented using concatenation and Network-in-Network (NiN) as proposed in [28, 34]. *Deconv* is realized as described in [37] using deconvolutions and L2 normalization. *Add* blocks use element-wise summation. Light red colored layers refer to ResNet-50s layers pretrained on ImageNet.

These recent fusion approaches explore the optimal stage to fuse streams during the feature extraction stage. Inspired by these methods we investigated different fusion strategies and block architectures. An overview of the resulting architectures is shown in Fig. 1.

**The Input Fusion:** This approach, first described for multispectral pedestrian detection by [52] is implemented by stacking the VIS and IR image channel-wise. Then, the resulting six-channel image is used as input to our model as shown in Fig. 1 (b). Additionally, the first convolution layer is adapted to process six-channel images by cloning the pretrained weights of the first convolution layer to get weights for the convolution of a six-channel image.

**Late Fusion Baseline:** We adapt the two separate feature extraction branches for the VIS and IR data to create three intermediate feature maps for both modalities. In contrast to other approaches, we do not fuse the streams during feature extraction and also do not require any additional fusion blocks as shown in Fig. 1 (c). Instead, we fuse both sets of feature maps implicitly in the detection head in the *conv_det*.

**Sparse Fusion:** Due to the similarities between our baseline architecture and a U-Net [44], we adapt the idea of *Sparse Fusion* proposed by Hazirbas et al. [20] for our multispectral CSPNet as visualized in Fig. 1 (d). Sparse Fusion was proposed within the authors FuseNet architecture, a commonly used U-Net based architecture that fuses RGB-D features for an image segmentation task. Following the authors, we continuously fuse the thermal feature maps into the color feature maps after each ResNet-50 layer using element-wise summation.

In addition to the previously described fusion approaches, we also investigate the performance of more complex fusion architectures that use dedicated fusion blocks that learn parameters to combine the features from both modalities.

**Proposed Halfway Fusion:** *Halfway Fusion* is defined as a fusion happening between the first and last convolutional layer of the feature extraction branch. On the contrary, fusion in the first convolutional layer is defined as *Early Fusion* and fusion after the last convolutional layer is defined as *Late Fusion* [28, 35, 52]. With ResNet-50 consisting of five convolutional layers [47], three different Halfway Fusion stages can be identified. We restrict our experiments to Halfway Fusion after *conv3_x* since [28] showed that a fusion at earlier stages performs significantly worse than other approaches. Based on the Halfway Fusion of König et al. [28], we fuse both streams after *conv3_x* and then proceed with the feature extraction on the fused features in a single branch. This is visualized in Fig. 1 (e).

**Proposed Late Fusion:** The *Late Fusion* as shown in Fig. 1 (f) combines the ideas of the previously described Halfway Fusion and the Late Fusion Baseline. Instead of a shared feature extraction branch proposed in the Halfway Fusion, we apply the idea of dedicated feature extraction streams for the IR and VIS features similar to the Late Fusion Baseline. However, we additionally incorporate the idea of fusion blocks from the Halfway Fusion to fuse each feature map separately with their corresponding feature map from the other modality and then upsample the fused product. This results in a much more fine-grained fusion of features, which is unique to our architecture.

## 3.3 Multispectral Data Augmentation

Current multispectral object detectors simply adopt the established data augmentations from VIS models and apply those identically to both the color VIS and the IR image. The baseline augmentations of CSPNet consist of *Random Flip*, *Random Rescale*, *Random Crop/Pave* and *ColorJitter* [37]. We remove the *ColorJitter* augmentation from the baseline augmentations as it can not be applied to monochromatic thermal IR images in a reasonable way. To the best of our knowledge, no researchers have focused on data augmentation specifically for multispectral images with the exception of Zhang et al. [55], who used a simple *RoI Jitter* augmentation to randomly jitter the box proposal's coordinates of one modality during training. However, this augmentation technique is exclusively limited to their model's realignment architecture and is not transferable to other models easily and intuitively.

We propose that augmentations for VIS and IR images should not be identical in a fusion architecture. Especially, since color VIS and thermal IR images show significantly different properties, an augmentation that is suitable for color VIS images might not have a similar effect on thermal IR images. Therefore, applying augmentations that are well-suited for only one of the two modalities and choosing a unique set of augmentations for each modality might improve the overall performance of a multispectral pedestrian detection model. At the same time this approach is universally applicable to all two-stream multispectral models. A general constraint to that approach is that one should avoid geometric transformations on the input images as those increase the misalignment between both modalities. This is problematic as established multispectral pedestrian benchmarks assume aligned images with only one GT bounding box per image pair. However, in misaligned cases, two separate GT annotations would be required to accurately assess a pedestrian's location.

To find suitable augmentations, we investigated various automated augmentation search approaches and augmentation surveys [6, 7, 21, 46, 58] as well as work on general thermal imagery [25, 26, 53] with the mentioned properties and constraints in mind. In this way, we identified two suitable augmentations *Random Erasing* and *Noise Injection* that seem to have some visible impact on the performance over the conventional augmentations. In addition, similar to Random Erasing, we propose a new *Random Masking* augmentation.

**Random Erasing:** Random Erasing has been shown to increase the performance of object detectors and to be compatible with our baseline augmentations [59]. Additionally, this augmentation can be applied synchronously to both input images, which means erasing rectangles at the same position in the VIS and IR image, as well as asynchronously, which means sampling different probabilities and parameters for the VIS and IR images separately while not affecting the misalignment. We use the parameter ranges provided by [59] from the person re-identification task.

**Proposed Random Masking:** Random Masking is a novel augmentation technique that we developed as an amplification of *Random Erasing*. Random Masking completely removes either one or none of the modalities with a probability of 0.5 by replacing the original image with an image where all pixel values are set to 0.

**Noise Augmentations:** Zhang et al. [58] showed that Gaussian noise models can be used to improve robustness of VIS based object detectors. However, for thermal IR images in real world applications, Poisson noise or salt-and-pepper noise are more common and suitable noise models [25, 26]. We investigate adding noise (with a probability of 0.2) to different modalities and report some plausible combinations in the ablation study in Table 2.

## 4  Experimental Results

**Implementation Details:** All our proposed models are implemented in the PyTorch framework [41]. The ResNet-50 backbone for both modalities is pretrained on ImageNet. We do not freeze the backbone during training to allow the net for adapting to the thermal IR modality. Unless stated otherwise, the model is trained on pedestrian instances that satisfy the All subset. We trained for 100 epochs with 2,000 image samples per epoch and a batch size of 12. Additionally, we apply a balanced sampling of images with and without pedestrians to counter the imbalance between images that contain pedestrians and those that do not. The used optimizer is Adam [27] with a learning rate of $1e^{-4}$. To reduce training time and allow bigger batch sizes during training, input images are downsampled to size $(384 \times 480)$ via the Random Crop/Pave and Random Rescale augmentations. During inference, we use a Non-Maximum Suppression (NMS) with an overlap threshold of 0.3, a confidence threshold of 0.01, and apply an aspect ratio of 0.41 to determine the width from the predicted scale.

**Dataset and Evaluation Metrics:** We report results on the challenging KAIST Multispectral Pedestrian Detection Benchmark using the established log-average Miss Rate (MR) over the range of $[10^{-2}, 10^0]$ False Positives Per Image (FPPI). We evaluate our models using the *Reasonable* and *All* subsets. Following the experimental setup provided by [24], the Reasonable subset consists only of pedestrians with at least 55 pixels of height and partial or none occlusion. To also evaluate our models for the aforementioned small-scale and highly occluded cases, we adopt the *All* setup of the Caltech Pedestrian Detection Benchmark [12, 13] (also denoted as *Overall* sometimes) for the KAIST benchmark allowing for pedestrians of any height and occlusion level.

**Impact of the Annotations:** Currently, three different training annotations for the KAIST Multispectral Pedestrian Detection Benchmark exist: the *original annotations* [24], the *sanitized annotations* [32], and the *paired annotations* [55]. Inaccuracies in the original labels motivated authors to do relabeling and thus improve the annotations of the dataset. However, Li et al. [32] only relabeled those images that contained Reasonable pedestrian instances according to the *original annotations* and therefore missed many small-scaled instances and heavily occluded instances. Zhang et al. [55] very recently published another set of annotations, called *paired annotations*, where they relabeled all annotations separately for the VIS and the IR modality. As the choice of the training annotations is not consistent in the literature, we provide an ablation study on the impact of these annotations on the per-formance. For that, we trained three *Late Fusion* models with identical settings on the three annotations and evaluated their performance under the KAIST All and Reasonable subset. To train our model on the *paired annotations* without biasing it towards one modality, we fuse the annotations of each modality by creating a new rectangular ground truth that exactly covers the union of the IR and VIS annotations. From the results in Fig. 2, we can see that the difference between *sanitized* and *paired annotations* on the Reasonable subset is just marginal, both significantly outperforming the *original annotations*. However, on the All subset the *paired annotations* significantly outperform the *sanitized annotations*. This gap on the All subset can be attributed to images containing only small-scale or heavily occluded instances that were relabeled in the *paired annotations* but not in the *sanitized annotations*. In order to be able to compare with the published state-of-the-art, we chose to use the *paired annotations* for a fair comparison for all of our experiments.

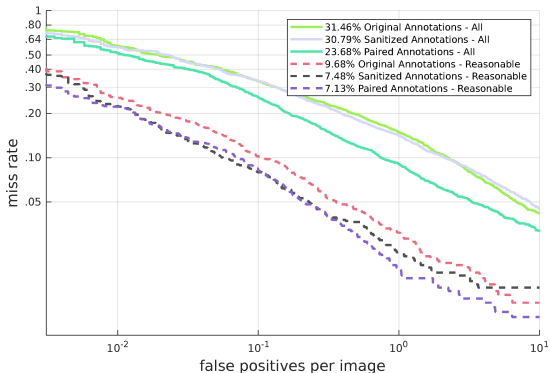

Figure 2: Effect of different annotations on evaluation under *Reasonable* and *All* settings.

| Architecture | Reasonable | | All | |
|---|---|---|---|---|
| | Average MR [%] | Min. MR [%] | Average MR [%] | Min. MR[%] |
| VIS Only | 20.50 ±0.75 | 20.18 | 43.06 ±0.55 | 42.80 |
| IR Only | 16.64 ±1.29 | 15.84 | 35.73 ±0.98 | 35.41 |
| Input Fusion | 10.14 ±0.48 | 9.68 | 30.44 ±0.74 | 29.76 |
| Late Fusion Baseline | 8.51 ±0.27 | 8.32 | 25.62 ±0.06 | 25.58 |
| Halfway Fusion | 7.55 ±0.57 | 7.21 | 23.93 ±1.02 | **22.87** |
| Late Fusion | **7.40** ±0.26 | **7.13** | 23.96 ±0.24 | 23.68 |
| Sparse Fusion | 8.30 ±0.57 | 7.79 | **23.88** ±0.15 | 23.71 |

Table 1: Results of the different architectures to fuse multispectral data. Bold results represent best log-average Miss Rate (MR).

**Impact of the Proposed Multispectral Fusion:** Following, we provide the evaluation of the different fusion strategies proposed in Section 3.1. In addition, we include results of the baseline reference DCNN individually trained on only the VIS or the IR modality denoted as *VIS Only* and *IR Only*. In Table 1, we can see that both Input Fusion and Late Fusion Baseline perform worse than other fusion architectures on both evaluation protocols. On the other hand, the gap between the Halfway Fusion and our proposed Late Fusion is rather small. Overall, however, Late Fusion performs best. While Sparse Fusion [20] marginally outperforms Late Fusion on the All subset, it fails to generalize on the Reasonable subset. Similarly on All, the Halfway Fusion achieves an insignificantly better MR over Late Fusion, which is evened by its slightly worse performance on the Reasonable subset. We decided to focus on the Late Fusion architecture for further experiments as it keeps a balanced MR on both evaluation protocols and its standard deviation is shown to be substantially smaller compared to the Halfway Fusion. The latter indicates a higher model robustness.

**Impact of Multispectral Data Augmentation:** We isolated the effect of each augmentation by adding only the augmentation technique listed in the *Augmentation* column on top of the baseline augmentations in our Late Fusion model. In order to compare these augmentations in Table 2, we can make the following observation: first, simple noise augmentation does not significantly improve the performance. However, the performance of noise augmentations is best when we sample separate probabilities for each noise augmentation indicated by the term *Async.*. While the differences are mostly small, this seems to indicate that using different noise augmentations for different modalities might be beneficial. Both Random Masking and Sync. Random Erasing lead to consistent improvements over our baseline augmentations. Additionally, using combinations of Random Masking and Random Erasing improve the miss rate even further on both subsets.

**Comparison with the State-of-the-Art:** As visualized in Fig. 3, we compare our detection method that is the proposed Late Fusion and Random Masking + Random Erasing Sync. on top of the baseline augmentations with five recent state-of-the-art approaches called Fusion RPN+BF [28], IAF R-CNN [33], IATDNN+IAMSS [19], MSDS-RCNN [32], and AR-CNN [55]. To fairly compare with the state-of-the-art, results on the Reasonable subset in Fig. 3 were obtained by only training on pedestrians with a height of more than 50 pixels as established by [32]. As clearly depicted, we outperform all existing models with a significant margin. Particularly, evaluating under the more challenging All subset, our method shows an improvement of more than 16 % over the best performing existing architecture. Furthermore, to show the effectiveness of our method especially for very small instances or heavily occluded instances, we plot the MR for specific pedestrian sizes and occlusion levels in Fig. 4. As seen, we clearly outperform the two current state-of-the-art models MSDS-RCNN [32] and AR-CNN [55]. The plots depicts that the performance gap is especially large for very small pedestrians with a height between 20 and 40 pixels ($\sim 40\%$ improvement over MSDS-RCNN and $\sim 25\%$ improvement over AR-CNN).

| Augmentation | Reasonable | | All | |
|---|---|---|---|---|
| | Average MR [%] | Min. MR [%] | Average MR [%] | Min. MR [%] |
| Baseline Augmentation | 7.40 ±0.26 | 7.13 | 23.96 ±0.24 | 23.68 |
| + Gaussian on VIS | ↑ 8.07 ±0.63 | 7.62 | ↑ 24.33 ±0.79 | 23.76 |
| + Gaussian on VIS, Gaussian on IR, Async. | ↑ 7.78 ±0.20 | 7.56 | ↓ 23.90 ±0.71 | 23.27 |
| + Gaussian on VIS, Poisson on IR, Async. | ↑ 7.77 ±0.64 | 7.25 | ↓ 23.95 ±0.90 | 22.92 |
| + Gaussian on VIS, Salt and Pepper on IR, Async. | ↑ 7.52 ±0.26 | 7.27 | ↓ 23.82 ±0.25 | 23.65 |
| + Random Erasing, Async. | ↑ 7.52 ±0.22 | 7.31 | ↑ 24.08 ±0.84 | 23.47 |
| + Random Erasing, Sync. | ↓ 7.29 ±0.33 | 7.03 | ↓ 23.58 ±0.54 | 23.1 |
| + Random Masking | ↓ 7.02 ±0.59 | 6.44 | ↓ 23.02 ±0.21 | 22.85 |
| + Random Masking + Random Erasing, Async. | ↓ **6.84** ±0.14 | 6.69 | ↓ 22.66 ±0.11 | 22.54 |
| + Random Masking + Random Erasing, Sync. | ↓ 6.96 ±0.48 | **6.42** | ↓ **22.29** ±0.33 | **21.95** |

Table 2: Comparison of different data augmentations. *Async.* means the augmentations were applied independently for both modalities, *Sync.* means the augmentations were applied identically on both modalities.

Finally, in Fig. 5 we provide a concise qualitative comparison between MSDS-RCNN [32], AR-CNN [55], and our approach, where we highlight small-scaled and highly occluded instances that the other competitive approaches do not detect confidently.

## 5 Conclusion

In this paper, we presented the first anchor box-free approach for multispectral pedestrian detection. By making a reference DCNN focus on pedestrian instances' center point and scale instead of the standard bounding box representation during training, we can create a model that performs particularly well for small-scaled and occluded pedestrians. Therefore, we adopted established multispectral DCNN fusion techniques and introduced them in a novel way to the reference network resulting in our Fusion CSPNet architecture. Within our Fusion CSPNet we then compared existing training annotations for the KAIST benchmark that provide valuable insight into their impact on a models performance. In addition, we proposed new multispectral data augmentation methods such as Random Masking that further improved the model. In this way, we outperform current state-of-the-art approaches by a large margin.

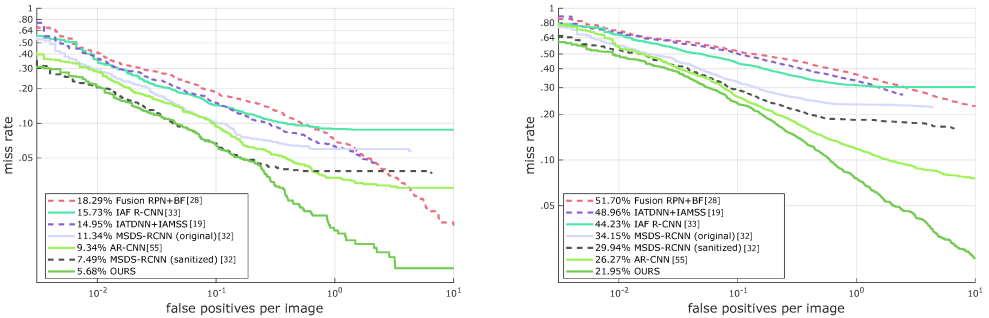

Figure 3: Miss rate plotted against False Positives Per Image (FPPI) for the KAIST Reasonable (left) and All (right) subset.

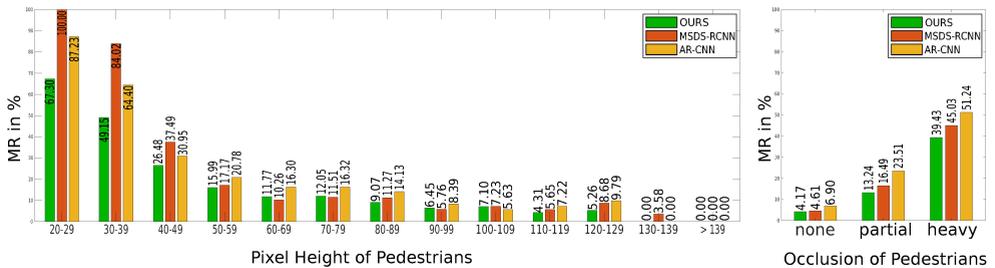

Figure 4: Distribution of log-average Miss Rates (MR) over different pedestrian sizes and different occlusion levels. Miss rates were taken on the interval of $[10^{-2}, 10^0]$ FPPI.

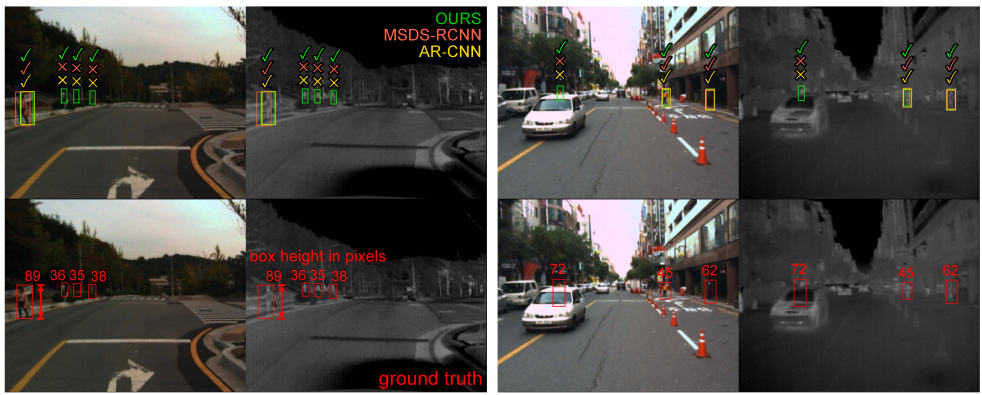

Figure 5: Comparison of detections taken from MSDS-RCNN [32], AR-CNN [55], and OURS in the top row. In the bottom row, we show corresponding ground truth boxes in red color. For better clarification we marked correctly detected pedestrians with a tick and missed pedestrians with a cross in the corresponding detectors color.

# Supplementary

## 6 Multispectral Data Augmentation

In Fig. 6 we provide a visualization of applying the *Random Erasing Async.* followed by the *Random Masking* data augmentations that were introduced in this paper.

## 7 Qualitative Comparison

In Fig. 7 we provide another set of examples, in which our detector is able to detect small-scaled instances where other approaches cannot distinguish between pedestrian or background correctly.

## 8 Quality of Annotations for the *All* Subset

In the context of evaluating under the challenging conditions of the *all* subset, we also did a qualitative investigation of the current test annotations. In Fig. 8 and 9 we show multiple instances, where our detector is able to find pedestrians that are not even labeled. Instead, these instances are only labeled in later frames as the pedestrian approaches the camera and increases in size. In the left column we show the frames, in which a pedestrian instance was not labeled but detected by a detector. In the right column we show the detections and annotations for one frame afterwards, where the pedestrian instance is labeled as it became larger. This brings us to the conclusion that, although our detector already outperforms many other approaches on the *all* subset, its performances would most likely improve even further, if these missing instances were labeled correctly. So, even though the dataset was already relabeled twice, there still is potential for improvement.

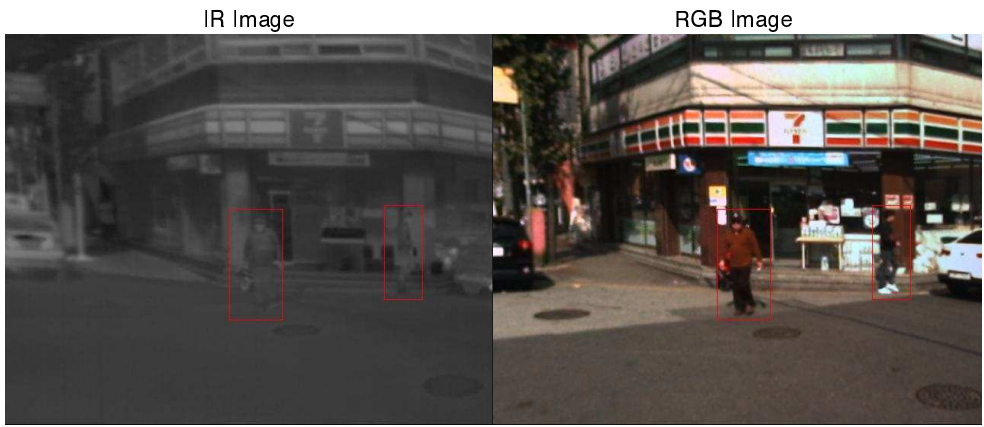

((a)) No augmentation

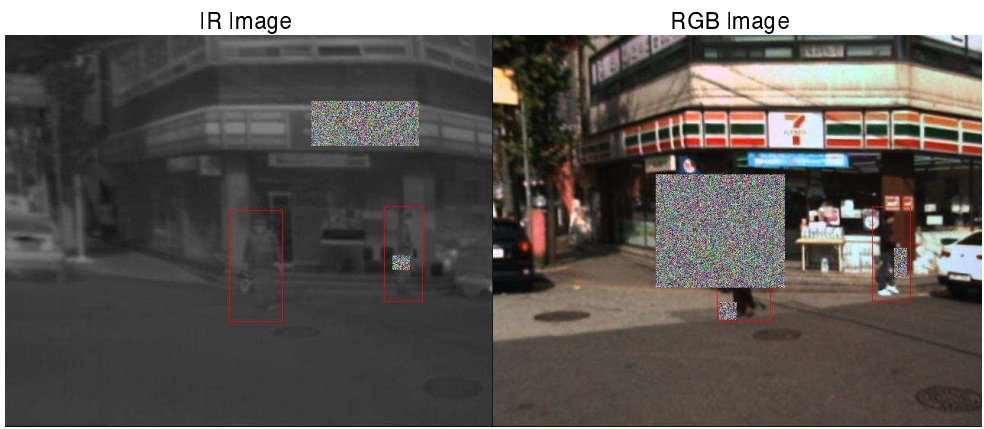

((b)) Random Erasing Async.

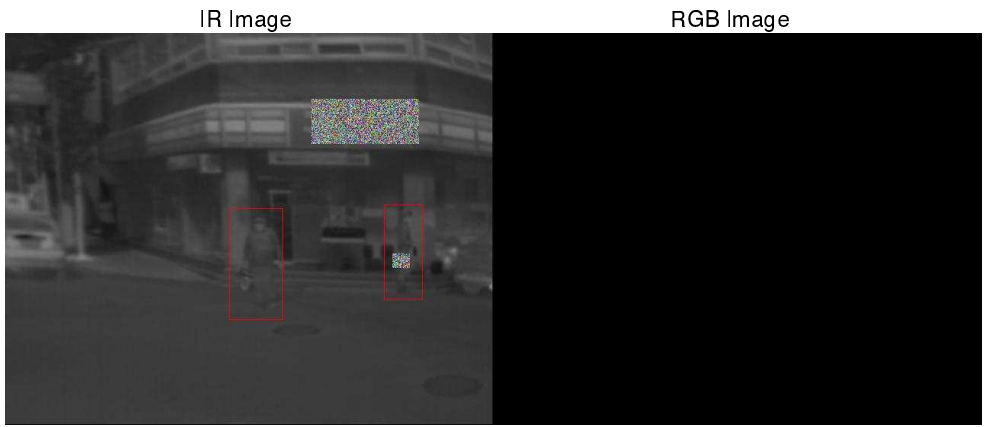

((c)) Random Masking

Figure 6: Exemplary visualization of *Random Erasing Async.* and *Random Masking* for an image pair from the KAIST dataset. Pedestrian GTs are marked with a red box for clarity.

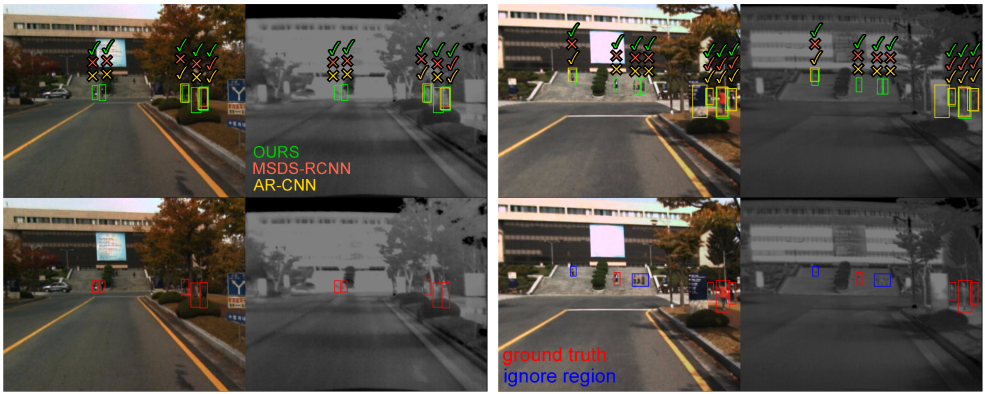

Figure 7: Comparison of detections of MSDS-RCNN [32] (orange), AR-CNN [55] (yellow) and OURS (green) in the top row. For better clarification we marked correctly detected pedestrians with a tick in the corresponding color of the detector and missed pedestrians with a cross in the corresponding color.

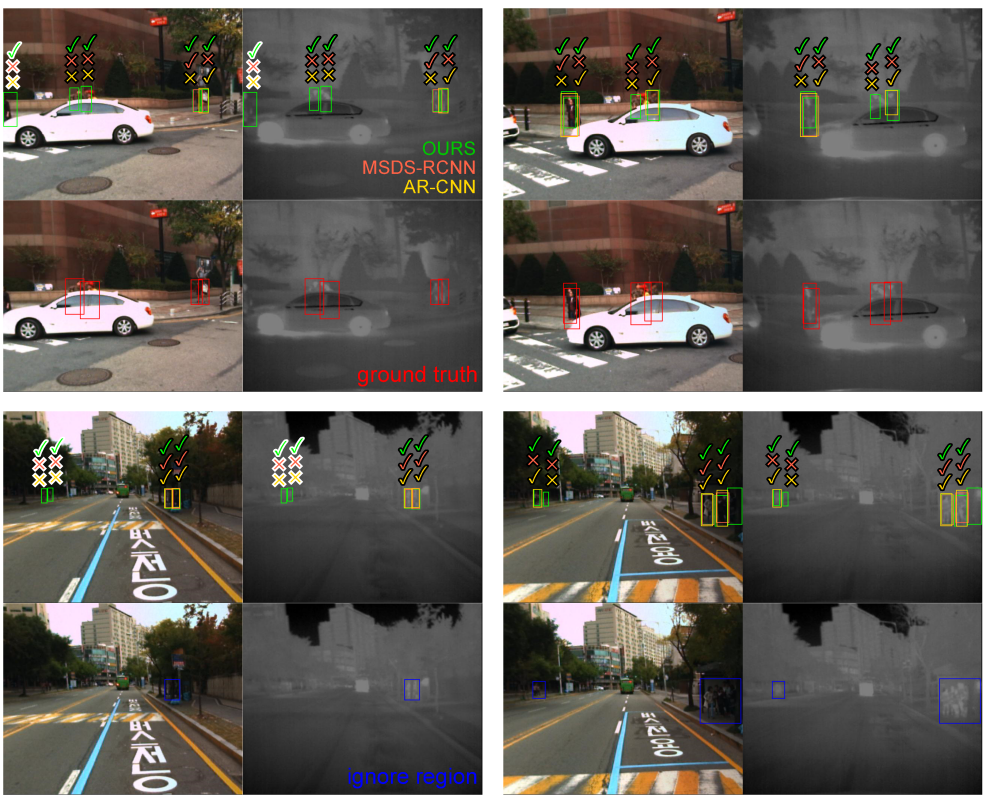

Figure 8: Multiple examples where our detector already detects pedestrians that are not labeled until one frame later. These cases are indicated by white framed ticks or crosses. Left column unlabeled instances, right column one frame later labeled.

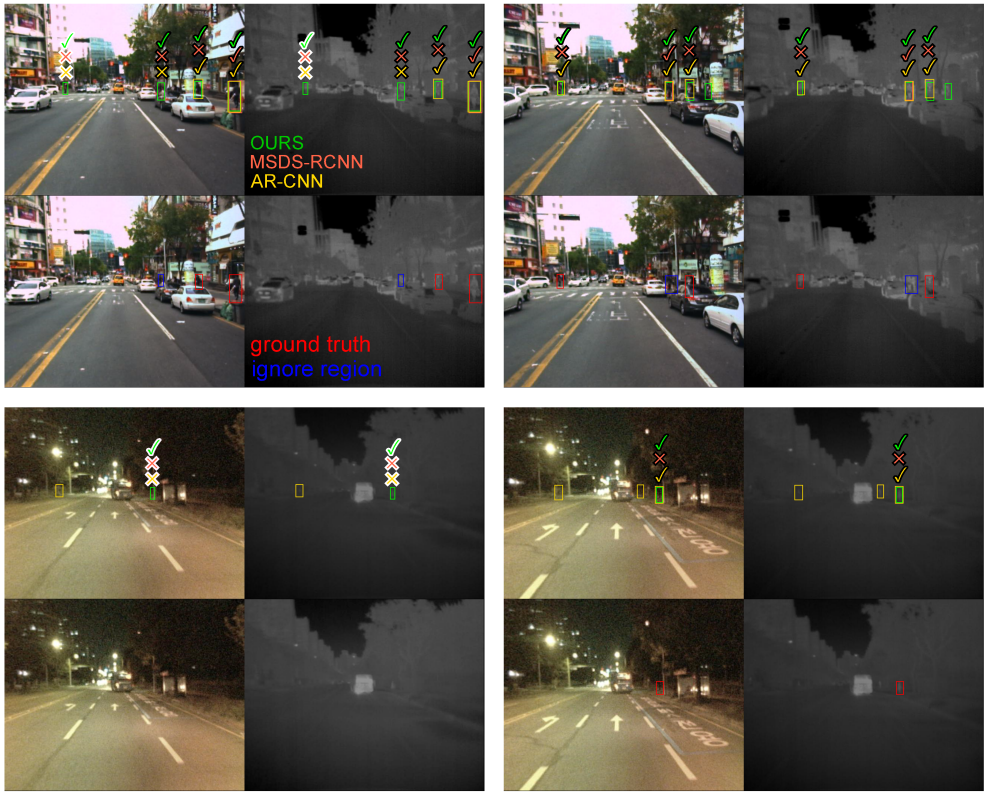

Figure 9: Multiple examples where our detector already detects pedestrians that are not labeled until one frame later. These cases are indicated by white framed ticks or crosses. Left column unlabeled instances, right column one frame later labeled.

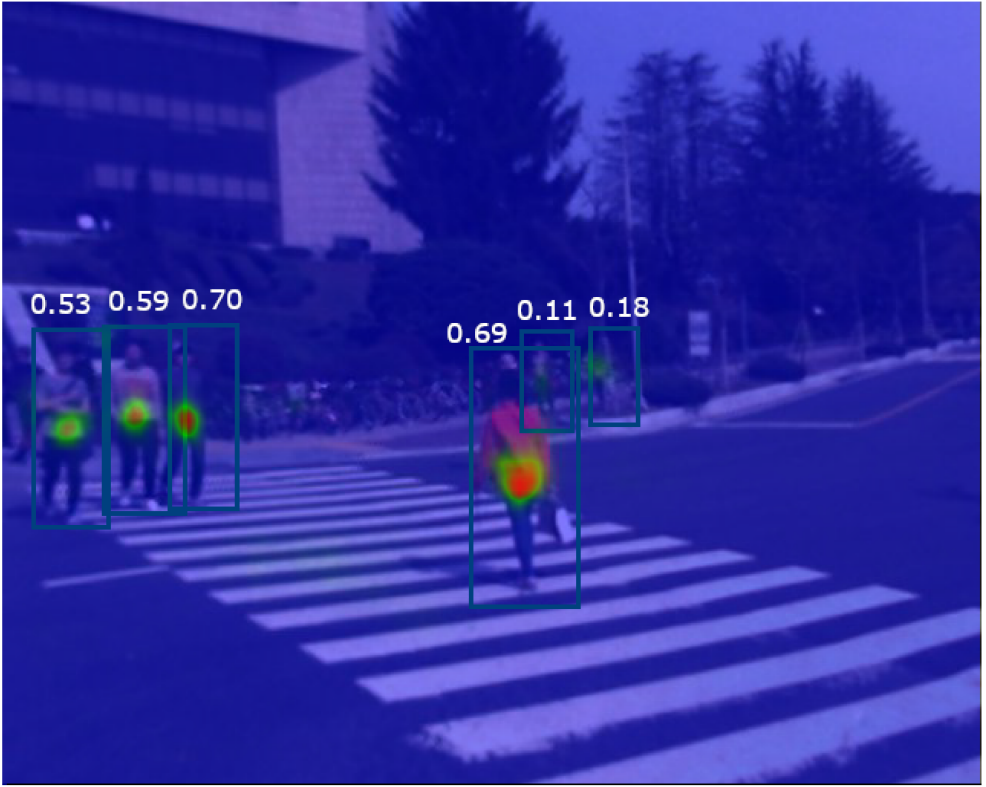

Figure 10: Visualization of the model's center map predictions as a heatmap. Detection confidence is depicted from blue (0.0) over green to red (1.0). The maximum confidence for each pedestrian instance is shown as white text. Ground truth annotations are visualized as green bounding boxes.